%% file: main.tex
\documentclass{article}

\usepackage[position, preprint]{neurips_2026}

\usepackage[utf8]{inputenc} 
\usepackage[T1]{fontenc} 
\usepackage{hyperref}  
\usepackage{url}   
\usepackage{booktabs}  
\usepackage{amsfonts}  
\usepackage{nicefrac}  
\usepackage{microtype}  
\usepackage{xcolor}   
\usepackage{amsmath}
\usepackage{amssymb}  

\usepackage{booktabs}
\usepackage{multirow}
\usepackage{longtable}
\usepackage{array}
\usepackage{ragged2e}
\usepackage{todonotes}
\usepackage{graphicx}
\usepackage{subcaption}

\usepackage{tcolorbox}
\usepackage{listings}
\tcbuselibrary{listings,breakable}
\usepackage{lipsum}

\newcolumntype{P}[1]{>{\RaggedRight\arraybackslash}p{#1}}

\title{Agentic AI Scientists Are Not Built For\\Autonomous Scientific Discovery}

\author{%
  \parbox{0.95\textwidth}{\centering
    \textbf{Harshit Bisht$^{1}$\quad
    Vinay Kumar$^{1}$\quad
    Kevin Maik~Jablonka$^{4,5,6}$\\
    Mausam$^{1,2}$\quad
    N.\,M.~Anoop~Krishnan$^{1,3}$}%
  }%
}

\input{hivemind/numbers}

\begin{document}

\maketitle

{\renewcommand{\thefootnote}{}%
 \footnotetext{%
   $^1$Yardi School of Artificial Intelligence,
   $^2$Dept.\ of Computer Science and Engineering,
   $^3$Dept.\ of Civil and Environmental Engineering,
   Indian Institute of Technology Delhi, India.
   $^4$Laboratory of Organic and Macromolecular Chemistry (IOMC),
   $^5$Center for Energy and Environmental Chemistry Jena,
   Friedrich Schiller University Jena, Germany.
   $^6$Helmholtz Institute for Polymers in Energy Applications Jena (HIPOLE Jena), Germany.
   \href{mailto:harshit.bisht@scai.iitd.ac.in}{\texttt{harshit.bisht@scai.iitd.ac.in}}, \href{mailto:vinay.kumar@scai.iitd.ac.in}{\texttt{vinay.kumar@scai.iitd.ac.in}}, \href{mailto:kevin.jablonka@uni-jena.de}{\texttt{kevin.jablonka@uni-jena.de}}, \href{mailto:mausam@iitd.ac.in}{\texttt{mausam@iitd.ac.in}},
   \href{mailto:krishnan@iitd.ac.in}{\texttt{krishnan@iitd.ac.in}}.}}

\begin{abstract}
A growing body of work pursues AI scientists capable of end-to-end autonomous scientific discovery. This position paper argues that although they already function as co-scientists, \textbf{agentic AI scientists are not built for autonomous scientific discovery}. We identify the following challenges in building and deploying autonomous AI scientists: (1) Problem selection is influenced by the McNamara fallacy; (2) Agents are built on large language models (LLMs) whose training corpora omit tacit procedural and failure knowledge of laboratory practice; (3) Preference optimisation during post-training compresses output diversity toward consensus; and (4) Most scientific benchmarks measure single-turn prediction accuracy and lack feedback from physical experiments back to the computational model. These challenges are not just questions of scale and scaffolding; they require revisiting fundamental design choices. To build truly autonomous AI scientists, we recommend scientific simulations as verifiers for training, a persistent, mutable epistemic state that carries beliefs and shifting objectives across an investigation, the establishment of a centralized preregistration repository for all AI-generated hypotheses, and application driven by scientific need rather than tool affordance.
\end{abstract}

\section{Introduction}
\label{sec:introduction}

\begin{quote}
\textit{``In general, we look for a new law by the following process.
First, we guess it; then we compute the consequences of the
guess and then we compare the result of the computation to
nature, with experiment or experience.
If it disagrees with experiment, it is wrong.
In that simple statement is the key to science.''}\\[4pt]
\hfill---\,Richard P. Feynman (1964)
\end{quote}

One of the ways science is defined is the interrogation of nature through a cycle of conjecture, prediction, and physical verification. What distinguishes scientific reasoning from other forms of structured reasoning is precisely this closing of the loop against experiment---the willingness to be proven wrong by nature. Artificial intelligence (AI) has begun to participate meaningfully in this enterprise~\citep{wangScientificDiscoveryAge2023,novikov2025alphaevolve0}. Across the natural sciences, AI systems have accelerated the processing of large experimental datasets~\citep{dagdelen_structured_2024}, made progress towards automated tool-use and laboratories~\citep{Darvish2025organa, aila}, uncovered latent structure in complex biological and chemical spaces~\citep{jumperHighlyAccurateProtein2021}, and reduced the time from hypothesis to computational candidate in domains ranging from drug discovery~\citep{coleyAutonomousDiscoveryChemical2020, coleyAutonomousDiscoveryChemical2020a} to materials design~\citep{zeniGenerativeModelInorganic2025,miret2025enabling,jain2013commentary,ahlawat_family_2026}.

Against this backdrop, a stronger proposition has gained traction: that large language model (LLM) based agentic systems can conduct natural science \emph{autonomously}. A growing body of systems now claims \emph{end-to-end autonomous discovery}---execution of a complete discovery cycle without meaningful human involvement: selecting problems worth pursuing, forming and revising hypotheses, designing and executing experiments, and communicating results~\citep{kitanoNobelTuringChallenge2021,gottweis2025towards,mitchenerKosmosAIScientist2025,luEndtoendAutomationAI2026,ghareebRobinMultiagentSystem2025,yamadaAIScientistv2WorkshopLevel2025,ghafarollahiSciAgentsAutomatingScientific2025,Boiko2023autonomous}. We refer to such systems as \emph{AI scientists}: LLMs deployed with the scaffolding of retrieval, tool use, critique, and memory. Major funding bodies are directing resources toward autonomous AI researchers~\citep{GenesisMissionNational2026, ARIA}; benchmark performance on scientific tasks is being cited as evidence of near-human scientific capability; and the Nobel Turing Challenge frames the aspirational endpoint as an AI system capable of foundational discovery in the natural sciences~\citep{kitanoNobelTuringChallenge2021}. If these claims are correct, the organisation of scientific research faces genuine transformation. If they are not, then the mismatch between the framing and the actual capability carries costs for how systems are evaluated and scientific effort is allocated.

Here, we argue that \textbf{current agentic systems are not built for autonomous natural science}---not because current systems lack scale or tooling, but because the shortcomings are \emph{inherent} to the training and deployment strategy. Unlike formal sciences such as mathematics where fast compilers and verifiers can address several of the shortcomings we identify, discovery in the natural sciences is bottlenecked by physical verification that is slow, partial, and itself requires interpretation~\citep{bungePhilosophyScienceVolume2017}; we centre our arguments there. We identify several challenges in building autonomous AI scientists, all the way from problem selection to application. The position advanced here is not that AI systems should be removed from scientific workflows---they are useful as a \emph{co-scientist}: a capable collaborator performing one or more tasks in the discovery process, while at least one of problem specification, operation of laboratory instrumentation, physical execution of experiments, and interpretation of results remains with human scientists~\citep{bianchiExploringUseAI2026}. This is exemplified by AI systems' success in problems that are well-stated by humans, primarily requiring optimisation in a well-defined problem space, and where constraints are defined a priori. Further, we do not claim that AI scientists are impossible, but argue that significant changes (Section~\ref{sec:agenda}) beyond scale and tooling are needed before they are feasible. The key contributions of the paper are as follows:
\begin{itemize}
 \item We identify challenges along four themes (detailed in Sections~\ref{sec:problem_selection}--~\ref{sec:evaluation}) that preclude autonomous scientific discovery: the McNamara fallacy in problem selection, the absence of tacit and failure knowledge from training corpora, diversity compression induced by preference optimisation, and the invalidity of current scientific benchmarks.
 \item We provide empirical evidence of cross-model convergence, consistent with diversity compression, through the \textit{hypothesis hivemind} experiment, showing that 12 frontier models from four providers, including open-weight models, converge semantically on both interpretive and open-ended hypothesis generation tasks in two natural-science domains.
 \item We outline several potential solutions to address these challenges and potentially realize AI scientists in Section~\ref{sec:agenda}.
\end{itemize}

\section{Challenges in autonomous discovery}
\label{sec:challenges}

The aspiration to automate scientific discovery predates modern AI by several decades \citep{lindsayDENDRALCaseStudy1993, langley1977bacon, king2009robot} 
. LLM-based agentic systems have renewed this ambition at a different scale. AI systems have proposed a drug candidate for dry age-related macular degeneration~\citep{ghareebRobinMultiagentSystem2025}, produced scientific reports spanning multiple domains in a single turn~\citep{mitchenerKosmosAIScientist2025,ghafarollahiSciAgentsAutomatingScientific2025}, and generated millions of candidate crystal structures, compressing what would have been decades of screening into days~\citep{merchantScalingDeepLearning2023,zeniGenerativeModelInorganic2025}. However, for each of these examples, human researchers still specified the problem, provided the dataset, and defined the criteria for success. We argue that this division of labour is not merely a short-term limitation fixable with scale but follows from design choices that do not fully reflect the spirit of scientific inquiry. The challenges we identify scale with the cost and completeness of the signal that tells a system it was wrong. In mathematics a proof assistant either verifies a result or returns a detailed error trace within seconds, and autonomous discovery is correspondingly more tractable; we do not dispute claims of autonomy in such domains. In materials discovery a density functional theory calculation takes hours and returns only a proxy for stability, while a synthesis attempt can take months and yields a result that must itself be interpreted before it counts as evidence.

Current systems that claim to be AI scientists are composed of a small and recurring set of components: retrieval over published literature; execution of tools and generated code; decomposition into multiple agents holding distinct roles, usually including a critic; iterative refinement of a candidate against that critic; memory or a shared notebook carried between calls; and, in a minority of systems, actuation of laboratory hardware~\citep{Boiko2023autonomous, Darvish2025organa, ghareebRobinMultiagentSystem2025, ghafarollahiSciAgentsAutomatingScientific2025, mitchenerKosmosAIScientist2025, luEndtoendAutomationAI2026}.

While this scaffolding improves these systems compared to a single LLM, the challenges we identify remain: (1) Problem selection happens before the scaffold is invoked; (2) Retrieval changes what the model can access but not what exists to be accessed, and thus cannot compensate for the tacit and failure knowledge gaps; (3) Critique and re-ranking by agents select among samples already drawn from the base model, reordering them without adding weight to regions that post-training has thinned (consistent with this, \citet{rios-garciaAIScientistsProduce2026} find the base model’s measured reasoning ability accounts for $41.4\%$ of explained variance in task success against $1.5\%$ for the scaffold, and conclude that scaffold engineering alone cannot repair the failures they observe); and (4) realistic evaluation and feedback from physical experiments pose challenges independent of AI scientist architecture. Table~\ref{tab:gap_analysis} outlines a summary of the specific challenges that any system aspiring toward autonomous discovery must solve. Sections~\ref{sec:problem_selection}--\ref{sec:evaluation} take each row in turn and examine why the gap exists in the present paradigm.

To ground our arguments, we discuss the search for solid-state battery electrolytes (SSE) with high ionic conductivity and electrochemical stability as a running example. SSE is one of the most impactful and well-studied problems at the frontier of energy storage research~\citep{dutraUnderstandingSolidstateBattery2025}. The atomistic mechanisms governing ion transport in solid electrolytes remain incompletely understood. No single readily computable objective captures all practical performance requirements. Progress demands coherent reasoning across scales---from quantum-mechanical density functional theory calculations, through molecular dynamics simulations capturing diffusive ion motion, to device-scale electrochemical models---each with its own intricacies. Finally, realizing the computationally designed electrolyte in a laboratory setting presents unique challenges including synthesizability, processability, and device-scale manufacturing. SSE discovery sits at the far end of the easy to hard verification spectrum described, and the complexity and urgency of the task make it a good backdrop to understand the challenges we outline.

\begin{table*}[t]
\centering
\footnotesize
\caption{The contrast between desired and current behaviour for the challenges in building AI scientists. Each row covers the challenges around a central theme analyzed in greater detail in Sections~\ref{sec:problem_selection}--\ref{sec:evaluation}. The solid-state electrolyte (SSE) discovery problem serves as the running case study.}
\label{tab:gap_analysis}
\setlength{\tabcolsep}{4pt}
\renewcommand{\arraystretch}{1.15}
\begin{tabular}{m{0.12\textwidth} m{0.25\textwidth} m{0.24\textwidth} m{0.32\textwidth}}
\toprule
\textbf{Challenge} &
\textbf{Desired behavior} &
\textbf{Current paradigm} &
\textbf{Gap, illustrated on SSE discovery} \\
\midrule

\textbf{Problem selection} & Identify questions that are field bottlenecks, integrating significance, tractability, and resource constraints & Select problems that are measurable, data-rich, and benchmark-amenable~\citep{haoArtificialIntelligenceTools2026} & Optimising ionic conductivity over interfacial stability, manufacturability, recyclability, and environmental impact is a human choice the system simply inherits \\

\addlinespace[1pt]

\textbf{Data bias} & Additional knowledge of which procedures work in practice and which approaches have been abandoned & Published literature, which records conclusions but omits tacit and failure knowledge~\citep{fanelliNegativeResultsAre2012} & Candidate rankings may omit unpublished knowledge of practical synthesis failures~\citep{merchantScalingDeepLearning2023} \\

\addlinespace[1pt]

\textbf{Diversity} &
Generate responses departing from prevailing consensus when the evidence warrants it & Compress outputs toward consensus through preference optimisation~\citep{kirkUnderstandingEffectsRLHF2023} & Convergence on familiar electrolyte families such as LLZO and NASICON rather than exploring under-represented or novel chemistries \\

\addlinespace[1pt]

\textbf{Benchmarking} &
Measure whether systems can weigh competing objectives and reason across extended investigations & Single-turn prediction accuracy on tasks vulnerable to leakage~\citep{kapoorLeakageReproducibilityCrisis2023} & High conductivity prediction scores are no indication of navigating the full discovery cycle from candidate proposal through synthesis and characterisation\\

\addlinespace[1pt]

\textbf{Inference \&} \textbf{physical} \textbf{validation} & Counterfactual reasoning, anomaly-driven belief revision, and feedback from physical results to computational models & Execute workflows but revise beliefs in only limited cases~\citep{rios-garciaAIScientistsProduce2026}; almost entirely in silico & A discrepant conductivity measurement should potentially revise the assumed transport mechanism; current systems discard this as noise or an outlier\\

\bottomrule
\end{tabular}

\end{table*}

\section{Problem selection: McNamara fallacy in AI-driven science}
\label{sec:problem_selection}

Problem selection is categorically different from problem solving; it involves judgment that integrates theoretical significance, practical tractability, community need, resource constraints, and the innate human need to understand the universe in ways that resist formal specification~\citep{nicklesScientificDiscoveryLogic1980,polanyiTacitDimension1966}. In Section~\ref{sec:challenges} we stated that current AI scientists rely on human judgement for problem selection; here we extend the argument by showing that the presence of AI technologies negatively impacts the human judgement they rely on.

The McNamara or Quantitative fallacy describes what happens when institutions face complex problems full of intangibles: They measure what is easy to measure, disregard what cannot be quantified, presume the unmeasured is unimportant, and finally declare that the unmeasured does not even exist~\citep{yankelovich1972corporate,omahony_medicine_2017}. AI techniques rely on numerical signals such as losses, rewards, or fitness values for optimization and evaluation of the final system, which naturally pushes research projects involving AI toward problems that can be quantified. In addition, machine learning techniques often require large amounts of data in a format that's easy to manipulate computationally, further reducing the span of problems amenable to AI solutions. This effect is already documented in scientific research. Across 41.3 million research papers, \citet{haoArtificialIntelligenceTools2026} found that AI-augmented scientists publish 3.02 times more papers and receive 4.84 times more citations, yet AI adoption shrinks the collective volume of scientific topics studied by 4.63\% and reduces scientist-to-scientist engagement by 22\%. The AI tools accelerate established fields leading to a narrower set of questions being asked, while genuinely hard, data-sparse problems go untouched.

Problem selection can be understood via the tension between \emph{technology-push} and \emph{demand-pull} innovation~\citep{distefanoTechnologyPushDemand2012}. In a technology-push regime, a highly-capable tool defines the agenda: researchers ask what can be done with it and search for amenable problems. In a demand-pull regime, researchers ask what is needed and build accordingly. The current AI deployment in science seems almost entirely technology-push. The SSE discovery problem illustrates where this leads. We have systems such as MatterGen~\citep{zeniGenerativeModelInorganic2025} which can generate candidate inorganic materials optimised toward any specified property constraint. The decision of which constraint to specify---ionic conductivity, interfacial stability, electrochemical window, or manufacturability in high-volume production---is not a conclusion the system reaches. Domain scientists must judge which bottleneck in solid-state battery science is currently most limiting and what experimental infrastructure is available to validate candidates before invoking AI tools. AI scientists must be trained to avoid relying on external problem selection mechanisms for performing autonomous discovery.

\section{Data bias: gaps in the training corpus}
\label{sec:pretraining}

Scientific publishing filters for some kinds of scientific knowledge over others by design or bias in the peer-review process. It selects for conclusions over process, for positive results over negative ones, and for ideas consistent with prevailing theory over those that challenge it. A model trained on this corpus does not inherit a snapshot of scientific knowledge but the output of that filter. Two types of omission matter most, and neither can be recovered by training on more of the same data.

\textbf{Tacit knowledge} Laboratory practice produces understanding that is rarely written down: which synthesis conditions are reliable, which reagents behave inconsistently across suppliers, which reported protocols require undocumented adjustments, and which measurements carry signatures of known experimental artifacts. \citet{polanyiTacitDimension1966} argued that this kind of knowledge is irreducibly non-propositional; an experienced researcher can recognise a result that looks too clean or a curve with the wrong shape, but cannot fully articulate the basis for that recognition that could be published or trained upon~\citep{fjellandWhyGeneralArtificial2020}. In SSE research, the tacit record includes which garnet-type synthesis routes consistently produce secondary phases under laboratory conditions that the published protocol does not acknowledge, which ionic conductivity measurements are artefacts of sample geometry rather than bulk material properties, and which candidate families have been screened and quietly set aside by multiple groups without a negative result ever appearing in print. A model trained on the published SSE literature sees the cleaned-up outputs of these decisions without any of the reasoning that produced them.

\textbf{Failure knowledge} Publication bias removes negative results from the corpus~\citep{fanelliNegativeResultsAre2012}. Further, the iterative cycle of anomaly identification, tentative re-framing, and targeted follow-up experimentation is not documented even when it eventually produces a publishable result. Models trained on the final paper have no access to these explored and abandoned paths through the hypothesis space. \citet{raccuglia2016machine} demonstrated that trials selected by a machine learning model trained on previously unpublished failed synthesis attempts outperformed candidates chosen by experienced chemists for predicting conditions for synthesising new vanadium selenite compounds. \citet{dingGenerativeAILacks2025} demonstrated another effect of missing failed tests in the training corpus: LLMs in simulated discovery settings could extend known results incrementally, but consistently failed to detect anomalies indicating that a prior hypothesis should be abandoned. We conclude that a corpus filtered for successful conclusions cannot teach how and when to abandon a hypothesis, an essential skill for autonomous discovery.

There is a third, more fundamental problem with training on a fixed corpus of publications. Statistical learning generalises to new samples from the training distribution. The most important results are out-of-distribution events in the space of scientific knowledge. A model trained on the published literature learns the consensus of the field as its prior, biasing it against disruptive suggestions that are too dissimilar to current consensus. This does not prevent cross-domain synthesis; LLMs can retrieve information in the literature of two disconnected fields and establish a connection across disciplinary boundaries similar to how \citet{shannonMathematicalTheoryCommunication1948} combined expertise in probability theory and electrical engineering to establish the study of information theory. However, the discovery of ideas in direct contradiction to the training corpus becomes unlikely. Shechtman observed a diffraction pattern with five-fold rotational symmetry in 1982, even though the established crystallographic restriction theorem classified it as a physical impossibility for periodic crystals. The finding was met with hostility by the community for several years; a statistical model would have likely discarded the finding for disagreeing with its training corpus. The anomaly was finally resolved by the existence of quasi-periodic crystals, and Shechtman received the Nobel prize in Chemistry 29 years after the original discovery~\citep{shechtmanQuasiPeriodicCrystalsLong2013}.
 
\section{Diversity: preference optimisation compresses the hypothesis space}
\label{sec:posttraining}

The drawbacks documented in Sections~\ref{sec:problem_selection} and~\ref{sec:pretraining} arise before a model generates a single output. A third challenge is introduced during post-training for alignment: reinforcement learning from human feedback (RLHF) and direct preference optimisation (DPO) adjust model outputs toward responses that are familiar and consistent with what the human annotators believe. \citet{kirkUnderstandingEffectsRLHF2023} showed these procedures measurably shrink the diversity of the output distribution toward annotator consensus. For scientific hypothesis generation, it is the wrong objective: the annotator preferences that shape the output distribution are products of the same consensus that the training corpus encodes. \citet{bellemare-pepinDivergentCreativityHumans2026} noted that state-of-the-art LLMs score below the mean of the top half of human participants on divergent association tasks measuring access to semantically remote concepts. This is the cross-domain reach that \citet{dunbarHowScientistsReally1995} identified as the primary driver of hypothesis revision in scientific practice. 

Even carefully designed post-training pipelines can get exposed to this bias through model distillation. \citet{cloudLanguageModelsTransmit2026} demonstrated that a teacher model's systematic tendency was faithfully acquired by a student trained solely on number sequences the teacher generated, with the bias invisible in any individual output. A model distilled from a preference-optimised teacher can inherit both the diversity compression and any additional biases the teacher developed even when they are not apparent in the training outputs. The problem can then compound across generations as AI-assisted literature increasingly populates future training corpora. As demonstrated by \citet{shumailovAIModelsCollapse2024}, training on AI-generated text degrades distributional tails and contracts output diversity across successive model generations, adding to the issue.

On the SSE task, the issues outlined in Sections~\ref{sec:pretraining} and~\ref{sec:posttraining} together raise the concern that AI systems asked to generate candidate hypotheses for improving lithium-ion conductivity in solid-state electrolytes may favour familiar electrolyte families, such as LLZO and NASICON, over underrepresented chemistries. These materials dominate the published literature and thus are highly represented in the model's priors. Scientific annotators for post-training are also likely to be familiar with the same literature which pushes the posterior after RLHF/DPO further towards them. These effects make it unlikely that AI systems trained this way will produce hypotheses that challenge the status quo.

\subsection*{Empirical evidence: the hypothesis hivemind}
\label{subsec:hivemind}
To test whether these theoretical predictions manifest in practice, we designed an experiment grounded in the dual-search framework of \citet{klahrDualSpaceSearch1988} and inspired by the \textit{Hivemind} hypothesis \citep{jiangArtificialHivemindOpenEnded2025}. We posed two tasks to 12 frontier models from four independent providers (Anthropic, Google, Moonshot AI, and OpenAI), four of them open-weight. We used 50 publications in the NeurIPS 2025 AI4Mat (AI for Accelerated Materials Discovery) track and 52 publications from PLOS Biology as the datasets (Appendix~\ref{app:dataset1} and~\ref{app:dataset2}). In the first task, models received a summary of experiments for a paper and were asked to recover the underlying hypothesis, an interpretive task with a determinate answer serving as a convergence baseline. In the second, models received the full paper text and were asked to generate new hypotheses to extend the work. We expected the open-ended nature of the second task to produce significantly more diverse outputs than the first. We generated 10 outputs per model for each paper and task and embedded them following the methodology of \citet{jiangArtificialHivemindOpenEnded2025}. Finally, we compared the cosine similarity of output embeddings within and across models and providers for both tasks.
 
\begin{figure}[t]
 \centering
 \includegraphics[width=\textwidth]{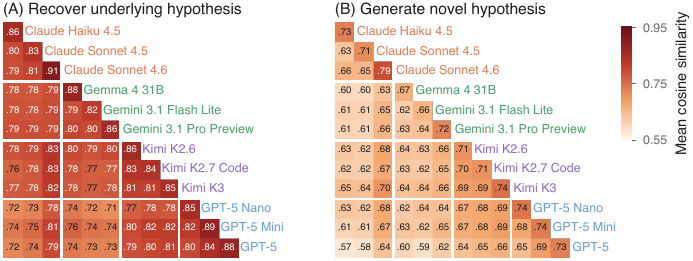}
 \caption{\textbf{Inter-provider output similarity remains consistently high, independent of the desired level of generative variety (AI4Mat; see Figure~\ref{fig:hivemind_plos_matrix} for PLOS Biology).} (A) Convergence desired: Heatmap of average cosine similarities of model output embeddings when asked to generate underlying hypotheses in response to experiment summaries. (B) Diversity desired: Heatmap of average cosine similarities of model output embeddings when asked to generate novel hypotheses in response to full publication text.}
 \label{fig:comparison}
\end{figure}

Figure~\ref{fig:comparison} (A) shows that similarity measures are high for the interpretive baseline, confirming that the embedding space detects semantic proximity when it is present. The critical result appears in Figure~\ref{fig:comparison} (B): inter-model similarities remain high despite the desired diversity in task outputs. Similarity is lower than for the baseline, but outputs from different providers are nearly as similar as repeated outputs from one model ($\hmAiFourmatNovelCrossProvider$ vs.\ $\hmAiFourmatNovelIntraModel$ on AI4Mat, $\hmPlosNovelCrossProvider$ vs.\ $\hmPlosNovelIntraModel$ on PLOS Biology). The effect is equally present for open-weight and proprietary models. Mean output length shows only a modest correlation with cross-model similarity (See Appendix~\ref{sec:hivemind} for details).

We further used the Vendi score~\citep{friedmanVendiScoreDiversity2023} to quantify the lack of diversity in responses. On the second task, querying ten hypotheses from ten different models instead of one led to an increase in effective number of distinct hypotheses from $\hmAiFourmatNovelIdeasOneModel$ to only $\hmAiFourmatNovelIdeasTenModels$ for the AI4Mat dataset. The corresponding increase for PLOS Biology was from $\hmPlosNovelIdeasOneModel$ to only $\hmPlosNovelIdeasTenModels$. Together, these measures capture that querying multiple AI systems for a scientific task might not lead to the desired variation for scientific hypotheses.

A research community that queries multiple AI systems for novel hypotheses gains only limited additional diversity over repeated sampling from a single model. For the running example of SSE, this raises the concern that consulting multiple frontier models when deciding which electrolyte candidates to synthesise may produce recommendations concentrated in the same families from each system independently. Synthesis and characterisation resources flow toward the region of candidate space the published literature already emphasises, while underrepresented chemistries and unconventional transport mechanisms receive little additional attention from the models queried.  Section~\ref{sec:evaluation} asks what happens when these accumulated failures meet the evaluation frameworks intended to detect them.

\section{Evaluation, inference, and physical validation}
\label{sec:evaluation}

Science is judged against what happens in physical reality. This section describes the problems when autonomous systems are evaluated, deployed, and applied to scientific research.

\subsection{What benchmarks measure}
\label{subsec:benchmarks}
\textbf{Construct validity} of a benchmark denotes whether high performance on a benchmark reflects the desired capability~\citep{cronbachConstructValidityPsychological1955,eriksson2025can}. \citet{asadiMIRAGEIllusionVisual2026} showed that fine-tuned models achieved top scores on chest X-ray VQA benchmarks without an image input. While the precise mechanism by which this happens is unknown, it is very likely that the examples selected for the benchmark had some language pattern exploited by the model, similar to the predictive value of information encoded in image backgrounds~\citep{xiao_noise_2020}. \citet{thaisAIScienceNeeds2026} document contradictory performance in \textsc{CORE-Bench Hard} and \textsc{ReplicationBench}, two methodologically similar benchmarks, implying that at least one of them does not sufficiently align with the desired capability. Data leakage is another concern. \citet{kapoorLeakageReproducibilityCrisis2023} surveyed 294 papers across 17 fields and found that correcting for eight common leakage types collapsed the apparent advantage of complex ML models over classical baselines.

\textbf{Ecological validity} represents the extent to which the tasks in a benchmark resemble real workflows~\citep{brunswikPerceptionRepresentativeDesign1956, eriksson2025can}. Scientific exploration is not single-turn; scientists explore many questions whose content depends on previous answers. Often, answers lead to the revision of the candidate list, experimental strategy, and the relative weight placed on different properties as information accumulates across weeks or months. Current benchmarks do not adequately capture this full structure. Many reward a correct prediction on a single turn rather than evaluate a sustained investigation which rewards a system for recognising that the prediction should alter the next question, or for maintaining a coherent investigative thread as objectives shift. This is further exemplified by \citet{rios-garciaAIScientistsProduce2026}'s evaluation across more than 25,000 agent runs and eight scientific domains. They found that evidence was ignored in 68\% of traces and refutation-driven belief revision occurred in only 26\% of cases.

\subsection{Reasoning across an investigation}
\label{subsec:inference}
The difference between scientific reasoning and exhibited model behaviour is more fundamental than benchmark validity. A working scientist treats counterfactual interrogation as a default. This disposition requires active cultivation~\citep{wasonFailureEliminateHypotheses1960}; humans themselves fail to apply it spontaneously without training. The operational core of \citet{popperLogicScientificDiscovery1959}'s falsificationism is the question a scientist asks of any result, including a favourable one: ``what would the evidence look like if my hypothesis were wrong?'' These questions are the trained default after years in a community of practice \citep{plattStrongInference1964}. Generating text conditioned on a counterfactual premise when prompted is a different operation. Currently, reasoning models supported by agentic harnesses do not reliably exhibit this disposition toward counterfactual interrogation~\citep{rios-garciaAIScientistsProduce2026}.

A scientific investigation also operates with shifting objectives. It is difficult to collapse this multi-objective structure into a single scalar reward, and reward specification is often seen as the central challenge in agentic AI~\citep{Jablonka2021,amodeiConcreteProblemsAI2016,kalinin2026bringing}. The appropriate reward function is not just difficult to specify a priori; it is non-stationary and changes in response to information that becomes available through the investigation. Taken together, we see that experiments that generate anomalous results as discussed in Sections~\ref{sec:pretraining} and~\ref{sec:posttraining} are likely to be discarded in a single-objective setting rather than treated as information for revising the reward signal.

Both points show up concretely in the SSE example. A researcher who obtains a computational prediction of high ionic conductivity does not proceed directly to synthesis; they check whether the prediction is an artefact of the simulation assumptions---the interatomic potential or the choice of density functional---whether the boundary conditions inflate bulk transport figures relative to a polycrystalline sample, and whether missing physical mechanisms, such as grain boundary resistance, secondary phase formation during sintering, or dendrite penetration at the electrolyte-anode interface, are likely to degrade performance. A system optimising a fixed scalar reward such as agreement between simulation and experiment might discard a candidate material that performs significantly worse in experiments than simulation instead of using it to improve its simulation methodology.

\subsection{From computation to physical reality}
\label{subsec:sim2real}
These failures in evaluation and inference set the conditions for the final gap: between what AI systems predict and what physical experiment confirms. The strongest demonstrations of AI scientific capability currently operate primarily in computational representations such as predicting crystal structures, proposing molecular candidates, and predicting protein structures~\citep{merchantScalingDeepLearning2023, jumperHighlyAccurateProtein2021}.

When \citet{cheethamArtificialIntelligenceDriving2024a} evaluated the crystal structures proposed by GNoME~\citep{merchantScalingDeepLearning2023} against novelty relative to known compounds, synthesizability with current technology, and practical utility for real applications, the overwhelming majority failed at least one criterion. Of the 2.2 million predicted structures, 736 had been independently experimentally realized at the time of publication. The intention is not to cite GNoME as an isolated example, but to highlight a pattern of generative models for materials that routinely omit feasibility constraints from their objectives. Across SSE discovery efforts, examples of computational candidates carried through to experimental validation at any scale remain scarce.

The in silico–in vitro performance gap cannot be closed by systems that treat steps of scientific discovery as independent prediction targets. The knowledge that bridges steps is relational and only becomes visible in the movement between them. Whether a predicted SSE composition is synthesizable from available precursors under achievable conditions relies on the tacit knowledge of synthesis practice described in Section~\ref{sec:pretraining}. Physical measurement adds a further layer of the same problem: an impedance spectrum does not arrive labelled, and reading off the separate contributions of bulk transport, grain boundary resistance, and electrode contact artefacts requires the kind of expertise that experienced researchers carry without writing down. Finally, physical results must feed back to revise the computational models that generated the predictions in the first place, a loop that current agentic systems have no mechanism to close.

\section{Alternative views}
\label{sec:alternative}

\textbf{\textit{Scale and scaffolding will close the autonomy gap.}} Improvements in context length, retrieval reliability, and tool-use accuracy will improve capabilities, but the failures described in this paper sit upstream of these properties (Section~\ref{sec:challenges}): more positive results cannot teach what failed experiments would, and a scaled-up model trained toward annotator consensus is likely to converge on the same region of hypothesis space at higher fluency. Experiments show that agentic scaffolding also fails to correct for this~\citep{rios-garciaAIScientistsProduce2026}.

\textbf{\textit{Demonstrated successes like AlphaFold and GNoME show that full autonomy is within reach.}} On the contrary, these contributions exhibit the boundary the paper is trying to draw. AlphaFold solved a problem formally stated for fifty years, with an unambiguous evaluation criterion available at experimental scale. GNoME searched a compositional space against a computable stability criterion. In both cases, human researchers specified the objective, defined the validation criterion, and interpreted what the result meant for subsequent work. Extrapolating these successes to autonomy assumes that this human judgment is unnecessary or reproducible by the same paradigm, which the evidence does not support. \citet{bianchiExploringUseAI2026} found that the highest-quality scientific outputs in a controlled study came from meaningful human-AI collaboration, with quality declining under full automation. The successes of the current paradigm are an argument for investing in the co-scientist model, not for replacing the human half of it.

\textbf{\textit{The distinction between co-scientist and autonomous scientist collapses as systems improve.}} If human oversight becomes increasingly nominal as capability grows, co-scientist might be a label applied to a workflow that is functionally autonomous. We argue that human involvement will not diminish randomly or uniformly; it will recede in precisely those aspects where the problems we identify are addressed, and these require different design commitments rather than more capability of the current kind. Until those are in place, the co-scientist model is an accurate description of what the collaboration requires: human judgment that supplements models where they are the most limited.

\textbf{\textit{Human scientists are also biased, limited, and frequently wrong.}} Individual human scientists are subject to confirmation bias, motivated reasoning, and the same consensus pressures that preference optimisation encodes \citep{wasonFailureEliminateHypotheses1960}. We believe the relevant comparison is between an AI system and the collective enterprise of science with corrective mechanisms such as peer review, replication requirements, adversarial collaboration, and reputational accountability for error developed over decades~\citep{longinoScienceSocialKnowledge1990}. A researcher who is wrong about a hypothesis faces replication attempts and critical commentary; but the convergence documented in Section~\ref{subsec:hivemind} means that querying additional AI systems is unlikely to apply the same corrective pressure.

\section{Toward autonomous AI scientists}
\label{sec:agenda}

Autonomous AI scientists nonetheless remain a goal worth pursuing. The limited availability of expert human judgement can constrain the rate of scientific progress. Further, long-horizon high-risk problems are attempted rarely because scientists are incentivised toward incremental projects~\citep{fosterTraditionInnovationScientists2015}. The scarcity of human judgement and risk-aversion can be reduced with highly capable AI scientists. We discuss some steps to make progress towards autonomous discovery.

\textbf{Public reviews for papers and proposals to train scientific judgment.} An autonomous scientific AI must be able to identify which questions are worth pursuing without explicit instruction. This requires a representation of scientific value that goes beyond what benchmarks can operationalise: an understanding of what constitutes a genuine field bottleneck, how theoretical significance and practical tractability trade off, and when a question is too well-answered to be worth pursuing further. The first step towards models capable of scientific judgement must be public data of it being exercised, such as open peer review~\citep{wolframOpenPeerReview2020}. Public reviews of research proposals will be even more impactful as records of evaluating scientific ideas without access to experiment results. We recognise that using such data for training involves legitimate concerns around reviewer consent, privacy, and copyright which must be addressed before such data is used in building systems.

\textbf{Public preregistration of AI-generated hypotheses and experiments.} Preregistration of hypotheses~\citep{nosekPreregistrationRevolution2018} before experimentation is mandatory in clinical trials, owing to the risk to public health and life from false positives \citep{angelisClinicalTrialRegistration2004, opensciencecollaborationEstimatingReproducibilityPsychological2015}. It has found less acceptance in other domains due to the additional work for research teams. We outline several reasons it is important, easy to implement, and impactful for the AI scientist paradigm.

Selective reporting and post-hoc analysis, colloquially described as ``p-hacking'' or ``post-diction'', are endemic to human science~\citep{ioannidisWhyMostPublished2005, simmonsFalsepositivePsychologyUndisclosed2011}. Agentic AI systems running automated hypothesis generation and testing loops at scale can amplify it by orders of magnitude, generating confirmatory results at a rate the peer review system was not designed to absorb. We propose that AI scientists be required to register hypotheses in a structured, machine-readable repository before any simulation or automated laboratory test is initiated. Unlike human researchers, this is easy to implement via an API call in the experimental pipeline, making circumvention a deliberate act against scientific integrity.

The repository can further serve as training data for two independent issues. The accumulated record of hypotheses and unsuccessful attempts to validate them will encode the failure knowledge described in Section~\ref{sec:pretraining}. The distribution of hypotheses generated by a community of AI scientists can also serve as a diversity and anti-redundancy training signal, creating a tractable objective for training the next generation of AI scientists to expand collective epistemic diversity. We acknowledge the challenges of generating consensus and compliance towards a centralized repository, but we believe the advantages we outline make it a worthwhile pursuit.

\textbf{Training corpora must encode what publishing discards.} The data bias documented in Section~\ref{sec:pretraining} cannot be corrected by training on more published literature. The tacit knowledge of laboratory practice and the failure knowledge accumulated through unpublished dead ends are absent from the published corpus by design, not by omission. Closing this gap requires 
different data sources. For tacit knowledge, the most promising direction is documenting the laboratory process such as video records of experimental procedures annotated by experienced researchers. This is expensive to generate and difficult to standardise, but it encodes knowledge that no text corpus contains. There have already been preliminary steps to build community resources in this direction such as~\citet{teytelmanProtocolsioVirtualCommunities2016}; further efforts to build open laboratory datasets of this kind would represent a qualitative shift in what training data for scientific AI can encode.

In addition to the suggestions of a public pre-registration dataset, training on failure knowledge requires changes in the incentive structure of scientific publishing. Journals that accept negative results, funding bodies that require deposition of failed experimental records, and community norms that treat documented failure as a scientific contribution rather than a career liability will help train better AI scientists. How much participation such a repository needs depends on how many failures precede a success in a domain and on the ratio at which failure data starts to help (Appendix~\ref{app:failure_data}). The AI community cannot generate this infrastructure alone, but it can make the case for why it is necessary, and build the tools needed to use it once it exists.

\textbf{Post-training that preserves diversity of ideas.} The convergence documented in Section~\ref{subsec:hivemind} is consistent with the diversity compression associated with preference optimisation. Recent work such as~\citet{chakrabortyMaxMinRLHFAlignmentDiverse2024, yaoNoPreferenceLeft2024} has attempted to modify the RLHF paradigm to have the post-trained model outputs match the full distribution of annotator responses. These techniques in combination with a large number of expert annotators can steer AI scientists towards generating diverse hypotheses.

\textbf{Simulators as scalable verifiers for scientific reasoning.}
The scientific community has built detailed simulation environments: molecular dynamics packages such as OpenMM~\citep{eastmanOpenMM8Molecular2024} and GROMACS~\citep{abrahamGROMACSHighPerformance2015} and physics engines such as MuJoCo~\citep{todorovMuJoCoPhysicsEngine2012} attempt to bridge the simulation-to-reality gap outlined earlier. These simulators are underexploited as training oracles for scientific AI, and can serve as sources of tacit and failure knowledge. Training on synthetic simulator data has been shown to improve performance on International Physics Olympiad problems with zero-shot transfer to real benchmarks~\citep{prabhudesai_solving_2026}. Simulators can generate varied RL training tasks beyond fixed published corpora~\citep{devvrit_art_2025}.

\textbf{Autonomous science requires a persistent epistemic state.} A context window alone does not provide a mechanism for revising beliefs and carrying them across an investigation. Scientific discovery requires an explicit, mutable representation of the agent's epistemic state that persists across the full arc of an investigation.

Shared memory or a read/write document, such as the structured world model to which \citet{mitchenerKosmosAIScientist2025} attribute their system's coherence over 200 agent rollouts, supports continuity across agent rollouts but does not by itself ensure explicit belief revision and uncertainty tracking. An adequate representation must also support introducing new hypotheses, updating degrees of belief, propagating uncertainty across dependent hypotheses, and selecting the next experiment by expected information gain. World models are one possible way to meet these requirements.

One possible implementation of epistemic state is a Bayesian network over hypotheses and observations where each node carries a posterior updated with experimental results, with new hypothesis nodes created in response to anomalies; drawing on concurrent graph data structures, multiple agents can update beliefs and insert nodes asynchronously. Meeting these requirements will likely demand integrating structured symbolic representations with learned neural ones---a non-trivial challenge the model-based RL community has begun to address in physical state spaces~\citep{haRecurrentWorldModels2018, hafnerMasteringDiverseControl2025}.

\section{Conclusion}
\label{sec:conclusion}

We establish that the gap towards autonomous scientific discovery has fundamental challenges that will not be closed by scale and scaffolding alone: the McNamara fallacy narrows problem selection, publication bias omits tacit and failure knowledge, preference optimisation compresses diversity, consistent with the convergence of 12 frontier models in the hypothesis hivemind experiment, and many benchmarks emphasise single-turn prediction over multi-step investigation and physical feedback. In SSE discovery, these challenges leave target choice to humans, may favour familiar electrolyte families such as LLZO and NASICON, and leave computational candidates far from battery devices. We suggest simulators as training verifiers, a persistent epistemic state across an investigation, and preregistration repositories as some of the potential ways forward.

\begin{ack}
This work was supported by the Yardi School of Artificial Intelligence, the Department of Computer Science and Engineering, and the Department of Civil and Environmental Engineering at IIT Delhi. The authors acknowledge the computational and storage resources provided by the IIT Delhi HPC facility and the Gemini Academic Program Award. H.B. and V.K. are supported by the ANRF IndiaAI PhD Fellowship and thank the NeurIPS 2026 Financial Assistance Award and the Yardi School of Artificial Intelligence for supporting their conference travel. The research of K.M.J. was supported by the Carl Zeiss Foundation, Open Philanthropy, Intel and Merck via the AWASES research center, a Google Research Scholar Award, and the NFDI consortium FAIRmat funded by the Deutsche Forschungsgemeinschaft (DFG, German Research Foundation) – project 460197019. Mausam is supported by grants from Tower Research, ANRF, Wipro, Verisk, and Google, as well as the Nick McKeown Chair Professorship. N.M.A.K. acknowledges funding support from a Google Research Scholar Award, the Anusandhan National Research Foundation (ANRF/ARG/2025/007405/ENS), and the Alexander von Humboldt Fellowship. Finally, the authors thank the reviewers for their insightful comments.
\end{ack}

\bibliographystyle{plainnat}
\bibliography{references}

\newpage
\appendix
\section{Experiment: hypothesis hivemind}
\label{sec:hivemind}

Since scientific hypothesis generation is highly open-ended, we were inspired by \citet{jiangArtificialHivemindOpenEnded2025} to evaluate the diversity of LLM outputs on scientific tasks by different providers. We chose 3 models each by Anthropic, OpenAI, Google, and Moonshot AI (Kimi) as highly popular AI model providers covering proprietary and open weight models, along with a task specialized model (Kimi K2.7 Code). Table~\ref{tab:hivemind_models} lists the models and the settings used to query them. Code, model outputs, and analysis scripts are available at \url{https://github.com/souspilot/hypothesis_hivemind}.

\begin{table}[!htbp]
 \centering
 \small
 \caption{Models used in the hypothesis hivemind experiment. Letters refer to the settings used for each dataset: the AI4Mat outputs of the Anthropic and OpenAI models were generated through the providers' own APIs, and all other outputs through OpenRouter.}
 \label{tab:hivemind_models}
 \input{hivemind/tables/models}
\end{table}

\subsection{Datasets and tasks}
\label{app:hivemind_setup}
We considered two datasets:
\begin{enumerate}
    \item \hmAiFourmatNPapers{} publications from the 2025 NeurIPS AI4Mat (AI for Accelerated Materials Discovery) track (full list in Appendix~\ref{app:dataset1}), chosen to be of shared interest to the AI and materials science communities. Paper text was extracted from the PDFs with the S2ORC parsing pipeline.
    \item \hmPlosNPapers{} publications from the April--June 2026 editions of PLOS Biology (full list in Appendix~\ref{app:dataset2}), chosen as an open access collection of papers from a different scientific domain. Paper text was taken from the publisher's JATS XML. A further \hmPlosNExcluded{} papers from the same issues were excluded because a provider content filter blocked every generation attempt for at least one model.
\end{enumerate}
We define 2 tasks:
\begin{enumerate}
 \item Recovering the underlying hypothesis given a summary of the experiments in a publication.
 \item Proposing novel hypotheses to extend the core findings given the full text of a publication.
\end{enumerate}
In the first task, models received a summary of experiments for each paper and were asked to recover the underlying hypothesis, an interpretive task with a determinate answer serving as a convergence baseline. The summaries were generated once per paper by Claude Sonnet 4.6, instructed to describe only the experiments and methods and to omit the abstract, results, and discussion, so that the hypothesis could not simply be copied. In the second, models received the title, abstract and full text of the paper and were asked to generate new hypotheses to extend the work. We expected the open-ended nature of the second task to produce significantly more diverse outputs than the first. The prompts used to generate model outputs can be found in Box~\ref{fig:prompts}.

\subsection{Generation and embedding}
\label{app:hivemind_generation}
For each paper, task, and model we drew \hmNSamples{} independent samples with identical prompts, giving $\hmNModels \times 102 \times \hmNSamples \times 2 = 24{,}480$ hypotheses in total. No sampling seed was fixed. Samples that were blank, returned an API error, or degenerated into repeated fragments (100 or more newlines; legitimate outputs contained at most 4) were discarded and regenerated until every model had \hmNSamples{} valid samples for every paper, so all comparisons below use equal sample sizes. All outputs were embedded with OpenAI's \texttt{text-embedding-3-small} similar to \citep{jiangArtificialHivemindOpenEnded2025}, and compared with cosine similarity.

\subsection{Similarity between and within models}
\label{app:hivemind_similarity}
\begin{figure}[!htbp]
 \centering
 \includegraphics[width=\textwidth]{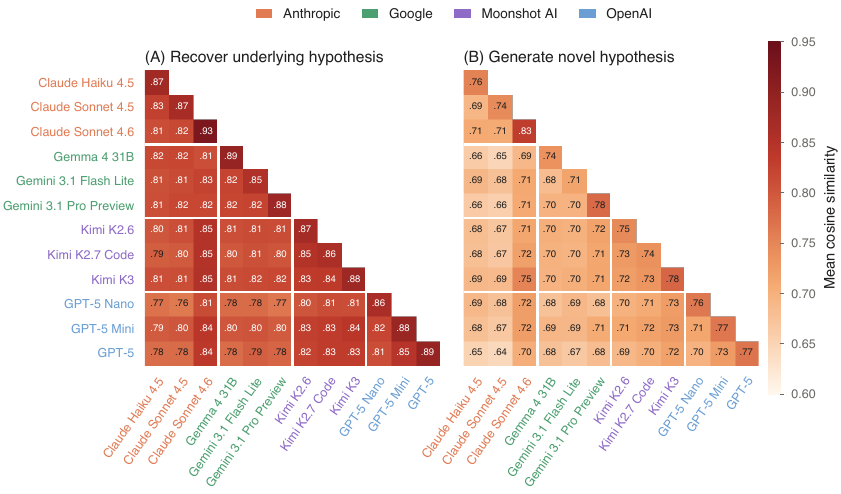}
 \caption{Inter-model similarities on PLOS Biology, the counterpart of Figure~\ref{fig:comparison}. (A) Recovering the underlying hypothesis. (B) Generating a novel hypothesis. Cells show the mean cosine similarity between the outputs of two models for the same paper; the diagonal compares a model with itself. Model names are coloured by provider.}
 \label{fig:hivemind_plos_matrix}
\end{figure}

\begin{figure}[!htbp]
 \centering
 \includegraphics[width=\textwidth]{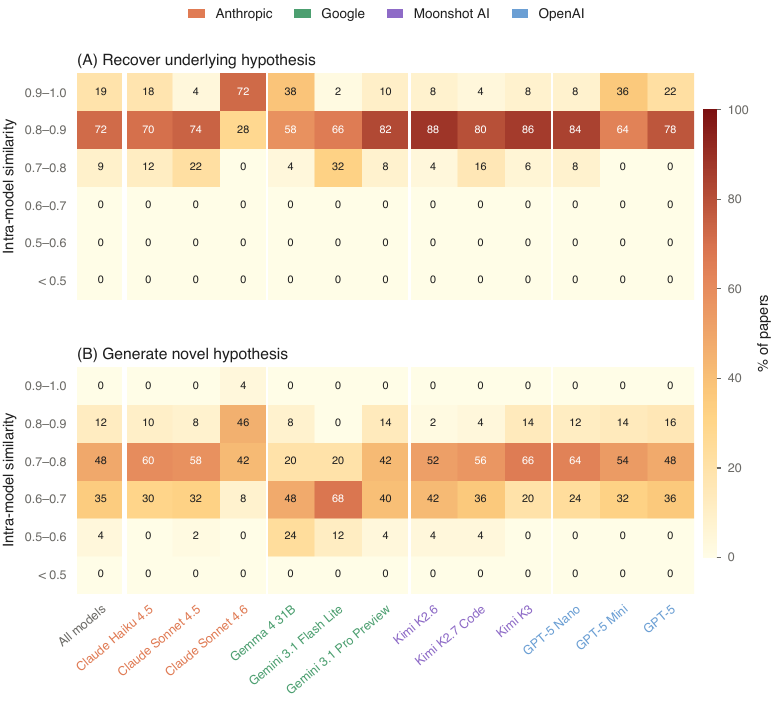}
 \caption{Intra-model similarities on AI4Mat for task 1 and task 2 as defined in the hypothesis hivemind experiment. Each column shows the percentage of papers for which a model's \hmNSamples{} samples have a given mean pairwise similarity; the first column pools all models. Together with Figure~\ref{fig:comparison}, we see that inter-model output similarities are not much lower than intra-model output similarities, which challenges the diversity in outputs produced by multiple LLMs collectively.}
 \label{fig:intramodel_similarity}
\end{figure}

\begin{figure}[!htbp]
 \centering
 \includegraphics[width=\textwidth]{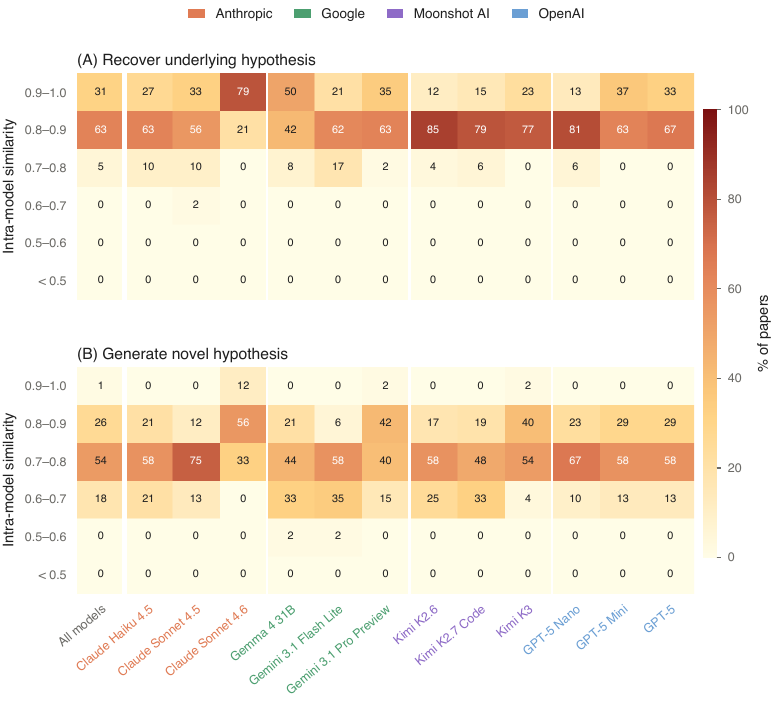}
 \caption{Intra-model similarities on PLOS Biology, as in Figure~\ref{fig:intramodel_similarity}.}
 \label{fig:intramodel_similarity_plos}
\end{figure}

We computed cosine similarity between intra-family and inter-family embedding groups, always between outputs for the same paper and excluding the similarity of a sample with itself. Each quantity is computed per paper and then averaged over papers:
\begin{itemize}
 \item \textbf{intra-model}: pairs of distinct samples from the same model;
 \item \textbf{intra-provider}: pairs of samples from two different models by the same provider;
 \item \textbf{cross-provider}: pairs of samples from models by different providers.
\end{itemize}
Table~\ref{tab:hivemind_summary} shows that the models' responses converged semantically for both tasks even though we only expected it for the first. For novel hypotheses on AI4Mat, samples from models by different providers are almost as similar to each other (\hmAiFourmatNovelCrossProvider) as samples from a single model are (\hmAiFourmatNovelIntraModel), and the same pattern holds on PLOS Biology (\hmPlosNovelCrossProvider{} against \hmPlosNovelIntraModel). Inter-model similarities are shown in Figure~\ref{fig:comparison} for AI4Mat and in Figure~\ref{fig:hivemind_plos_matrix} for PLOS Biology, while intra-model similarities are shown in Figures~\ref{fig:intramodel_similarity} and~\ref{fig:intramodel_similarity_plos}.

\begin{table}[!htbp]
 \centering
 \small
 \caption{Mean cosine similarity between hypotheses for the same paper, with 95\% bootstrap confidence intervals over papers.}
 \label{tab:hivemind_summary}
 \resizebox{\textwidth}{!}{\input{hivemind/tables/similarity_summary}}
\end{table}

\subsection{Embedding sanity check}
\label{app:hivemind_sanity}
\begin{figure}[!htbp]
 \centering
 \includegraphics[width=\textwidth]{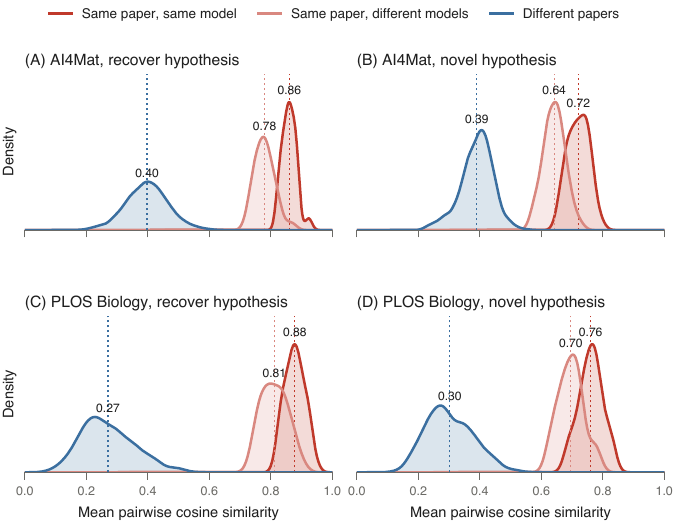}
 \caption{Cosine similarity distribution of embeddings of outputs by any model when generated in response to the same or different papers. The different distributions illustrate that the similarity observed across different model-outputs reflects a lack of semantic diversity, not degenerate behaviour of the embedding model. Dashed lines and labels mark the means.}
 \label{fig:distribution}
\end{figure}

To verify that the embedding model was not showing degenerate behaviour by embedding all the outputs close to each other (since they were all from the scientific domain), we looked at the cosine similarities for embeddings of outputs generated with the same or different papers as input. Figure~\ref{fig:distribution} shows that the embedding model is capable of distinguishing semantic differences: outputs for different papers have a mean similarity of \hmAiFourmatNovelLevelDifferentPaper{} (AI4Mat) and \hmPlosNovelLevelDifferentPaper{} (PLOS Biology) for novel hypotheses, far below the \hmAiFourmatNovelLevelCrossModel{} and \hmPlosNovelLevelCrossModel{} between different models answering for the same paper.

\subsection{Effective diversity}
\label{app:hivemind_diversity}
\begin{figure}[!htbp]
 \centering
 \includegraphics[width=\textwidth]{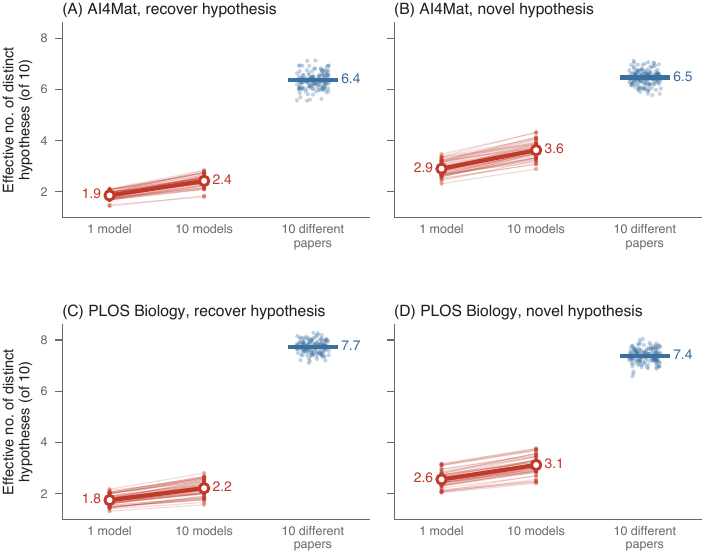}
 \caption{Effective number of distinct hypotheses (Vendi score) among \hmNSamples{} hypotheses. Each thin line is one paper: \hmNSamples{} samples from one model (left) and one sample each from \hmNSamples{} different models (right); thick lines are the means. The blue cloud shows \hmNSamples{} hypotheses about \hmNSamples{} different papers, for scale.}
 \label{fig:hivemind_diversity}
\end{figure}

Mean pairwise similarity does not directly say how many different ideas a set of hypotheses contains. We therefore also measured the Vendi score \citep{friedmanVendiScoreDiversity2023}, the exponential of the entropy of the eigenvalues of the cosine-similarity matrix normalised by the number of samples, which can be read as the effective number of distinct items in the set. We compare sets of exactly \hmNSamples{} hypotheses so that set size cannot drive the result: a single model's \hmNSamples{} samples for a paper; one sample each from \hmNSamples{} randomly chosen models for the same paper (200 draws per paper); and, as a scale reference, \hmNSamples{} hypotheses about \hmNSamples{} different papers. For novel hypotheses on AI4Mat, drawing from \hmNSamples{} models instead of one raises the effective number of distinct hypotheses only from \hmAiFourmatNovelIdeasOneModel{} to \hmAiFourmatNovelIdeasTenModels{} (95\% bootstrap CIs over papers [2.8, 3.0] and [3.5, 3.7]), against \hmAiFourmatNovelIdeasTenPapers{} for hypotheses about different papers; on PLOS Biology the values are \hmPlosNovelIdeasOneModel{}, \hmPlosNovelIdeasTenModels{}, and \hmPlosNovelIdeasTenPapers{} (Figure~\ref{fig:hivemind_diversity}). The increase is present for every paper, but small.

\subsection{Output length}
\label{app:hivemind_length}
Models differ in how long their hypotheses are (Table~\ref{tab:hivemind_length}), which could affect embedding similarity. Across the \hmNModels{} models, mean output length correlates with intra-model similarity for novel hypotheses ($r = \hmAiFourmatNovelLenCorrIntra$ on AI4Mat, $r = \hmPlosNovelLenCorrIntra$ on PLOS Biology), but only weakly with a model's similarity to the other models ($r = \hmAiFourmatNovelLenCorrCross$ and $r = \hmPlosNovelLenCorrCross$). Output length therefore does not account for the cross-model convergence we observe.

\begin{table}[!htbp]
 \centering
 \small
 \caption{Mean (standard deviation) output length in characters, and Pearson correlation across models between mean length and mean similarity.}
 \label{tab:hivemind_length}
 \input{hivemind/tables/output_length}
\end{table}

\clearpage
\subsection{Prompts}
\label{app:hivemind_prompts}
\begin{tcolorbox}[
 title=Prompts used in experiments,
 colback=gray!5,
 colframe=black,
]
\label{fig:prompts}
\textbf{Task 1: Generating experiment summaries}
\begin{lstlisting}
SYSTEM_PROMPT = (
 "You are a helpful assistant for summarizing key details of experiments and methodologies from scientific papers."
)
USER_INSTRUCTION = (
 "Summarize the following research paper, focusing ONLY on this question:
 Carefully analyze ONLY the experiments performed or methods used.
 Do NOT include results, abstract, introduction, or discussion.
 Output MUST be valid JSON of the form:
 `{'
 ` "title": "<paper title>",'
 ` "experiments_summary": "<concise summary>"'
 `}'
 Do NOT wrap the JSON in markdown code fences.
 Paper text:
)
\end{lstlisting}

\textbf{Task 1: Generating underlying hypothesis}
\begin{lstlisting}
SYSTEM_PROMPT = (
 "You are a scientific reasoning assistant. Given a description of the experiments and methods from a research paper, infer the underlying hypothesis being tested - the core scientific claim the experiments were designed to validate. A hypothesis is a specific, testable, and falsifiable prediction about the relationship between variables. Output ONLY the hypothesis as a single declarative sentence. Do not include preamble, explanation, or any other text."
)
USER_INSTRUCTION = (
 "Generate a single testable hypothesis based on the experiment description above. Express it as one declarative sentence (e.g. `If X, then Y because Z')."
)
\end{lstlisting}

\textbf{Task 2: Generating novel hypothesis}
\begin{lstlisting}
SYSTEM_PROMPT = (
 "You are an expert research scientist. Given the context of a research paper, your task is to generate a single novel hypothesis that logically extends beyond the paper's existing findings - not a restatement of them. The hypothesis must be: (1) grounded in a gap or open question identified in the paper, (2) specific and testable, (3) falsifiable. Output ONLY the hypothesis as a single declarative sentence with no preamble or explanation."
)
USER_INSTRUCTION = (
 "Based on the research context above, generate one novel hypothesis that extends beyond what this paper has already established."
)
\end{lstlisting}
\end{tcolorbox}

\subsection{Dataset 1: AI4Mat}
\label{app:dataset1}
The following papers were considered for the hypothesis hivemind experiment. They are selected from the list of accepted papers from the AI4Mat (AI for Accelerated Materials Discovery) workshop at NeurIPS 2025.

\begin{enumerate}
 \item Data-driven prediction of polymer surface adhesion using high-throughput MD and hybrid network models \\ \url{https://openreview.net/pdf?id=0SPoKR8Xrk}
 \item STR-Bamba: Multimodal Molecular Textual Representation Encoder-Decoder Foundation Model \\ \url{https://openreview.net/pdf?id=0uWNuJ1xtz}
 \item Universally Converging Representations of Matter Across Scientific Foundation Models \\ \url{https://openreview.net/pdf?id=12ZCZVKm7r}
 \item Preference Learning from Physics-Based Feedback: Tuning Language Models to Design BCC/B2 Superalloys \\ \url{https://openreview.net/pdf?id=24lzMGlvnq}
 \item Pharmacophore-Guided Generative Design of Novel Drug-Like Molecules \\ \url{https://openreview.net/pdf?id=35aDuh7ndX}
 \item Cross Modal Predictive Architecture for Material Property Prediction \\ \url{https://openreview.net/pdf?id=3WZkuWlzmN}
 \item Language Model Enabled Structure Prediction from Infrared Spectra of Mixtures \\ \url{https://openreview.net/pdf?id=3pAVbjWMXW}
 \item ML-Driven Discovery of Metastable States \\ \url{https://openreview.net/pdf?id=4U2k4uw43B}
 \item Universal Machine Learning Interatomic Potentials Enable Accurate Metal-Organic Framework Molecular Modeling \\ \url{https://openreview.net/pdf?id=4Xh9oL5rH0}
 \item GEOM-Drugs Revisited: Toward More Chemically Accurate Benchmarks for 3D Molecule Generation \\ \url{https://openreview.net/pdf?id=57YLCp7n2V}
 \item Generalizable Prediction of Mixture Etching Rates Using Graph Neural Networks \\ \url{https://openreview.net/pdf?id=5OsnDm1CdX}
 \item Foundation Models Enabling Multi-Scale Battery Materials Discovery: From Molecules to Devices \\ \url{https://openreview.net/pdf?id=6pjxodugzO}
 \item GAP: Guided Diffusion for A Priori Transition State Sampling \\ \url{https://openreview.net/pdf?id=7brF4sMQq3}
 \item Task Alignment Outweighs Framework Choice in Scientific LLM Agents \\ \url{https://openreview.net/pdf?id=7cbwuA5k0T}
 \item A Computational Workflow for Cost-Effective Synthesis of Inorganic Materials: Integrating Thermodynamics, Cellular Automata, Machine Learning, and Commercial Databases \\ \url{https://openreview.net/pdf?id=7l75CbxtmC}
 \item MatPROV: A Provenance Graph Dataset of Material Synthesis Extracted from Scientific Literature \\ \url{https://openreview.net/pdf?id=8JFITrNy3K}
 \item Accelerated Inorganic Materials Design with Generative AI Agents \\ \url{https://openreview.net/pdf?id=9JSO4qf1RQ}
 \item The Loss Landscape of XRD-Based Structure Optimization Is Too Rough for Gradient Descent \\ \url{https://openreview.net/pdf?id=A21WF9M1Um}
 \item AI-Guided Design and Discovery of Silicon-Based Anode Materials for Lithium-Ion Batteries \\ \url{https://openreview.net/pdf?id=AQkGpEMGWA}
 \item LLM Agents for Knowledge Discovery in Atomic Layer Processing \\ \url{https://openreview.net/pdf?id=Bg4Hn9Qq3w}
 \item Towards Dynamic Benchmarks for Autonomous Materials Discovery \\ \url{https://openreview.net/pdf?id=Cfj7uBu5dy}
 \item Surrogate Modeling for the Design of Optimal Lattice Structures using Tensor Completion \\ \url{https://openreview.net/pdf?id=Ciw6DbDa4U}
 \item Scalable Low-Energy Molecular Conformer Generation with Quantum Mechanical Accuracy \\ \url{https://openreview.net/pdf?id=Ei3eF8B8XH}
 \item Towards End-to-End Learning of Protein Structure Prediction and Structure-based Sequence Design \\ \url{https://openreview.net/pdf?id=EuACaJblk4}
 \item Training Speedups via Batching for Geometric Learning: An Analysis of Static and Dynamic Algorithms \\ \url{https://openreview.net/pdf?id=Gzf8k2wPdF}
 \item Closed-loop, Machine Learning-Driven Optimization of Reactor Yields in Reactive Carbon Electrolyzers \\ \url{https://openreview.net/pdf?id=InZczCC8X1}
 \item Benchmarking Agentic Systems in Automated Scientific Information Extraction with ChemX \\ \url{https://openreview.net/pdf?id=YKxwBMK8Nl}
 \item Accurate Band Gap Prediction in Porous Materials using $\Delta$-Learning \\ \url{https://openreview.net/pdf?id=a3LKICpDO2}
 \item Accelerated Discovery of High-Performance Polyamines for Solid-State Direct CO$_2$ Capture via Efficient Simulations and Bayesian Optimization \\ \url{https://openreview.net/pdf?id=aECXy5Jgm4}
 \item Efficient Nudged Elastic Band Method using Neural Network Bayesian Algorithm Execution \\ \url{https://openreview.net/pdf?id=acfR6umMJt}
 \item A Chemically Grounded Evaluation Framework for Generative Models in Materials Discovery \\ \url{https://openreview.net/pdf?id=amn6lBDjXm}
 \item NaviDiv: A Comprehensive Tool for Monitoring Chemical Diversity in Generative Molecular Design \\ \url{https://openreview.net/pdf?id=auRe7zr32I}
 \item When Forces Disagree: A Data-Free Fast Uncertainty Estimate for Direct-Force Pre-trained Neural Network Potentials \\ \url{https://openreview.net/pdf?id=bmgU7yWBeC}
 \item AMDEN: Amorphous Materials DEnoising Network \\ \url{https://openreview.net/pdf?id=cEgjPFdLvl}
 \item CHROMA: Conversational Human-Readable Optical Multilayer Assembly for Natural Language-Driven Inverse Design of Structural Coloration \\ \url{https://openreview.net/pdf?id=cFTvHHXvt6}
 \item Coupling Language Models with Physics-based Simulation for Synthesis of Inorganic Materials \\ \url{https://openreview.net/pdf?id=ctyy8EJYQj}
 \item An Effective Machine Learning Frame for Materials Discovery Structured by a Chemical Concept \\ \url{https://openreview.net/pdf?id=dEtRvi7G5i}
 \item Accelerating Material Discovery for Metal Organic Frameworks using Large Language Models \\ \url{https://openreview.net/pdf?id=dmeAH1hVR8}
 \item Concept-based Steering of Large Language Models for Conditional Molecular Generation \\ \url{https://openreview.net/pdf?id=e8bcQehZ15}
 \item An Exploration of Dataset Bias in Single-Step Retrosynthesis Prediction \\ \url{https://openreview.net/pdf?id=eUiZg9uUt4}
 \item Benchmarking Knowledge Transfer Methods in De Novo Materials Discovery \\ \url{https://openreview.net/pdf?id=egi8g2U0ZX}
 \item Towards Fully Automated Molecular Simulations: Multi-Agent Framework for Simulation Setup and Force Field Extraction \\ \url{https://openreview.net/pdf?id=enQdbinvNd}
 \item SAM-EM: Real-Time Segmentation for Automated Liquid Phase Transmission Electron Microscopy \\ \url{https://openreview.net/pdf?id=farKrjdsIH}
 \item CompGen: A Conditional Generation Framework for Inverse Composition Design of Catalytic Surfaces \\ \url{https://openreview.net/pdf?id=g6Sj1OFjAu}
 \item Physics-Constrained Diffusion for Lightweight Composite Material Design \\ \url{https://openreview.net/pdf?id=gifMFKvAl5}
 \item XDIP: A Curated X-ray Absorption Spectrum Dataset for Iron-Containing Proteins \\ \url{https://openreview.net/pdf?id=hFzjgQzoVU}
 \item Machine Learning Interatomic Potentials: Library for Efficient Training, Model Development and Simulation of Molecular Systems \\ \url{https://openreview.net/pdf?id=hQCdhenqre}
 \item Semi-Supervised Learning for Molecular Graphs via Ensemble Consensus \\ \url{https://openreview.net/pdf?id=hk6iX4mg3B}
 \item Emergent Pose-Invariance in 3D Molecular Representations via Multimodal Learning \\ \url{https://openreview.net/pdf?id=iFHaZzs6Kz}
 \item Bridging Data-Rich and Data-Poor Domains on Lithium-Ion Battery via Scanning Electron Microscopic Data Through Convolutional Neural Network Transfer Learning \\ \url{https://openreview.net/pdf?id=j3aOU8Ahue}
\end{enumerate}

\subsection{Dataset 2: PLOS Biology}
\label{app:dataset2}
The following papers were considered for the hypothesis hivemind experiment. They are selected from the April--June 2026 issues of PLOS Biology; \hmPlosNExcluded{} further papers from these issues were excluded as described in Appendix~\ref{app:hivemind_setup}.

\begin{enumerate}
  \item Structural insights into subtype-specific agonist recognition by sphingosine-1-phosphate receptors\\
        \url{https://doi.org/10.1371/journal.pbio.3003381}
  \item Structured water molecules drive activation and G protein selectivity in the GPR174 receptor\\
        \url{https://doi.org/10.1371/journal.pbio.3003447}
  \item Circadian rhythms remain temperature compensated during a Q neuron--induced hibernation-like state in mice\\
        \url{https://doi.org/10.1371/journal.pbio.3003475}
  \item Cryptococcus neoformans adapts to host CO2 concentrations via metabolic and stress-response remodeling\\
        \url{https://doi.org/10.1371/journal.pbio.3003561}
  \item Incorporating variation in death times improves predictions of ectotherm responses to stressful temperatures\\
        \url{https://doi.org/10.1371/journal.pbio.3003623}
  \item A cortico-subthalamic circuit rapidly engages and releases inhibition of specific movements depending on the environmental context\\
        \url{https://doi.org/10.1371/journal.pbio.3003635}
  \item Expansion of the geranylgeranyl pyrophosphate synthase gene family underlies the evolution of terpenoid biosynthesis in termites\\
        \url{https://doi.org/10.1371/journal.pbio.3003648}
  \item Emotion regulation success involves systematic gradient-based reconfigurations of large-scale activation patterns in the human brain\\
        \url{https://doi.org/10.1371/journal.pbio.3003666}
  \item Neural signaling contributes to heart formation and growth in the invertebrate chordate, Ciona robusta\\
        \url{https://doi.org/10.1371/journal.pbio.3003715}
  \item Capsular specificity in temperate phages of Klebsiella pneumoniae is driven by diverse receptor-binding enzymes\\
        \url{https://doi.org/10.1371/journal.pbio.3003716}
  \item Glucose is dynamically regulated by time of day in humans and Drosophila\\
        \url{https://doi.org/10.1371/journal.pbio.3003717}
  \item Macrophages self-generate and refine chemotactic gradients during migration towards complement C5a\\
        \url{https://doi.org/10.1371/journal.pbio.3003728}
  \item Casein kinase 2-mediated phosphorylation of the splicing factor SF3B3 plays a key role in esophageal squamous cell carcinoma progression\\
        \url{https://doi.org/10.1371/journal.pbio.3003729}
  \item Diversity patterns in terrestrial tetrapod clades are governed by equilibrium dynamics\\
        \url{https://doi.org/10.1371/journal.pbio.3003730}
  \item The non-vesicular cholesterol transporter GRAMD1C is a pan-coronavirus antiviral target\\
        \url{https://doi.org/10.1371/journal.pbio.3003736}
  \item The m6Am methyltransferase PCIF1 promotes osteogenic differentiation of mesenchymal stem cells through stabilization of Wnt-related transcripts\\
        \url{https://doi.org/10.1371/journal.pbio.3003739}
  \item Sleep strengthens successor representations of learned sequences in humans\\
        \url{https://doi.org/10.1371/journal.pbio.3003740}
  \item Wild parrots exhibit age-dependent conformity when learning about novel food\\
        \url{https://doi.org/10.1371/journal.pbio.3003741}
  \item Bacteria producing the bioplastic polyhydroxybutyrate kill the nematode Caenorhabditis elegans\\
        \url{https://doi.org/10.1371/journal.pbio.3003748}
  \item Somatosensory input drives membrane potential dynamics in motor cortex during voluntary limb movement\\
        \url{https://doi.org/10.1371/journal.pbio.3003749}
  \item A dual-threshold system relying on multiple c-di-GMP metabolic enzymes controls cell fate of a cyanobacterium\\
        \url{https://doi.org/10.1371/journal.pbio.3003750}
  \item Aberrant dopaminergic activity during consolidation causes age-related memory generalization in Drosophila\\
        \url{https://doi.org/10.1371/journal.pbio.3003752}
  \item Action information is integrated into entorhinal representations of conceptual space and is reflected in eye movements\\
        \url{https://doi.org/10.1371/journal.pbio.3003755}
  \item Human brains construct individualized global rankings from identical few-shot learning input\\
        \url{https://doi.org/10.1371/journal.pbio.3003756}
  \item Neural population dynamics and temporal context cells in macaque medial parietal cortex support temporal order memory\\
        \url{https://doi.org/10.1371/journal.pbio.3003759}
  \item Genomic and functional adaptations in the guanylate-binding protein GBP5 highlight specificities of bat antiviral innate immunity\\
        \url{https://doi.org/10.1371/journal.pbio.3003760}
  \item Dopamine D2-receptor blockade in humans disrupts the effect of effort on learning\\
        \url{https://doi.org/10.1371/journal.pbio.3003765}
  \item Ultrasound neuromodulation reveals distinct roles of the dorsal anterior cingulate cortex and anterior insula in learning\\
        \url{https://doi.org/10.1371/journal.pbio.3003767}
  \item Emergent spatial structure in the gut microbiota is driven by bacterial growth and gut contractions\\
        \url{https://doi.org/10.1371/journal.pbio.3003772}
  \item Polyphosphate modulates the stress-responsive formation of functional RNA-protein condensates in bacteria and mammalian cells\\
        \url{https://doi.org/10.1371/journal.pbio.3003775}
  \item Structure of the human P2X3 receptor reveals the basis for subtype-selective inhibition by sivopixant\\
        \url{https://doi.org/10.1371/journal.pbio.3003777}
  \item Past and present goals are represented concurrently during visual search\\
        \url{https://doi.org/10.1371/journal.pbio.3003779}
  \item The KMT2F histone methyltransferase interacts with the RNA polymerase I machinery to promote ribosomal RNA transcription\\
        \url{https://doi.org/10.1371/journal.pbio.3003785}
  \item Acoustic delivery of indocyanine green via biosynthetic gas vesicles for tumor photothermal therapy\\
        \url{https://doi.org/10.1371/journal.pbio.3003786}
  \item Encoding performance of cortical neurons critically depends on their morphological and neurophysiological properties\\
        \url{https://doi.org/10.1371/journal.pbio.3003789}
  \item Environmental uncertainty shapes human effort learning\\
        \url{https://doi.org/10.1371/journal.pbio.3003791}
  \item Mannose-binding lectin 2 secreted by hepatocellular carcinoma cells recruits and activates natural killer cells to reshape an immune-activated microenvironment\\
        \url{https://doi.org/10.1371/journal.pbio.3003793}
  \item USP39 promotes antiviral defense through post-transcriptional control of RIG-I and stabilization of STING\\
        \url{https://doi.org/10.1371/journal.pbio.3003796}
  \item The anti-neural role of BMP signaling is a consequence of its ancestral function in dorsoventral patterning\\
        \url{https://doi.org/10.1371/journal.pbio.3003803}
  \item Environmental redox conditions and strain variation define phenazine-mediated antagonism in co-infecting bacteria\\
        \url{https://doi.org/10.1371/journal.pbio.3003809}
  \item Olfactory bulb-cortex oscillations encode perceived odor intensity rather than concentration\\
        \url{https://doi.org/10.1371/journal.pbio.3003810}
  \item Sleep deprivation increases levels of the synaptic density marker SV2A in the human brain\\
        \url{https://doi.org/10.1371/journal.pbio.3003816}
  \item Splicing deficiency is driven by genomic erosion in non-recombining algal mating-type chromosomes\\
        \url{https://doi.org/10.1371/journal.pbio.3003823}
  \item Flexible goal learning involves coordinated population activity in dCA1 and medial orbitofrontal cortex\\
        \url{https://doi.org/10.1371/journal.pbio.3003824}
  \item Phenotypic heterogeneity optimizes trade-offs during adaptive deployment of the type VI secretion system\\
        \url{https://doi.org/10.1371/journal.pbio.3003838}
  \item Metabolic organization of macaque visual cortex reflects visual field topography and perceptual specialization\\
        \url{https://doi.org/10.1371/journal.pbio.3003847}
  \item Multiple adhesion molecules act together in oligodendrocyte-mediated axonal selection and myelin formation\\
        \url{https://doi.org/10.1371/journal.pbio.3003854}
  \item Growth arrest of Mycobacterium tuberculosis in acidic environments enhances their survival of antibiotic treatment\\
        \url{https://doi.org/10.1371/journal.pbio.3003857}
  \item Argonaute 2 drives resistance to immune checkpoint inhibitors in immunorefractory non-small cell lung cancer\\
        \url{https://doi.org/10.1371/journal.pbio.3003860}
  \item Learning engages transient and sustained cellular mechanisms in the human brain\\
        \url{https://doi.org/10.1371/journal.pbio.3003861}
  \item Normative assembly rule reveals fairness in microbial communities\\
        \url{https://doi.org/10.1371/journal.pbio.3003872}
  \item Distinct sources of decision-related signals in visual cortex are represented in different local field potential bands\\
        \url{https://doi.org/10.1371/journal.pbio.3003873}
\end{enumerate}

\section{How much failure data is enough?}
\label{app:failure_data}

The participation a repository of failure knowledge would need to meaningfully improve AI scientists cannot be stated as a single figure: it depends on the domain, scope, and complexity of the problems selected, and establishing it is a research question in its own right. Results from the literature instead identify which factors determine such a figure.

\citet{striethkalthoffMachineLearningChemical2022} studied why models trained on the published chemical literature predict experimental outcomes so poorly. Drawing between 37k and 190k published reactions per reaction type across four types from the Reaxys database, they could not train a useful predictor on any of them, obtaining correlation coefficients below 0.5 and average errors of 15\% or more, against errors of roughly 5\% for models trained on systematically varied experimental data. They tested whether this was due to bias in dataset construction, where similar reactions are reported repeatedly: in one reaction type, almost 45,000 of the 75,600 recorded reactions used a single reagent, and reproducing that concentration in the training data raised prediction error by about 30\%. Imposing the published distribution of outcomes, which favours successes over failures, reduced the correlation coefficient from 0.78 to 0.49, and did so more severely the smaller the training set. Prediction errors improved substantially with only a few thousand varied examples, showing that a small number of failed or nearly failed examples can carry more information than 100k or more similar ones.

\citet{toniatoNegativeChemicalData2025} used a transformer pretrained on 396,145 reactions drawn from US patents as their baseline, which reached 52.9\% accuracy on the target reaction family before any further training. Applying reinforcement learning to that model using 22 successful reactions and 748 synthetically generated negative examples, they reached 58.6\% accuracy, whereas conventional fine-tuning on the same 22 successes yielded 54.9\%, which the authors report as no improvement over the starting model. In a less scarce setting, with 220 successful examples against the same 748 negative examples, the ordering reversed and fine-tuning outperformed reinforcement learning. Failure records are therefore most valuable when successful trials are scarce, and in their setting the transition lies between success-to-failure ratios of 1:3.4 and 1:34. The result of \citet{raccuglia2016machine}, discussed in Section~\ref{sec:pretraining}, points the same way: archived failures from a single laboratory were enough to outperform experienced chemists.

\citet{yinQuantifyingDynamicsFailure2019} studied how failures are distributed across science and entrepreneurship, comparing grant applications that succeed after iterative revision with those that never do. Their reported success rates offer an estimate of how many successful and failed attempts each participating community member could contribute.

The required participation is therefore a function of two quantities: how many unsuccessful hypotheses precede a successful one in a given domain, and the critical ratio beyond which learning from a mixture of successes and failures outperforms learning from successes alone.

\section{AI scientists and the scientific community}
\label{app:community}

Forecasting how a true AI scientist would reshape the scientific community is difficult, and the discussion below is correspondingly speculative.

\subsection{Why autonomy is worth pursuing}
Our primary motivation is to counter claims of autonomous discovery already being made in the community, rather than to argue that agentic AI scientists are or are not the field's highest priority. We nonetheless believe that autonomous scientists would enable progress beyond what AI co-scientists can achieve. Unlike computation, which can be parallelised, human judgement must be applied sequentially, which makes it a binding constraint on the rate of scientific progress. High-risk problems whose cycles from hypothesis to experimental confirmation extend across long time horizons are attempted rarely, not because they are unimportant, but because scientists are strongly incentivised toward conservative, incremental projects that ensure steady productivity over risky exploration~\citep{fosterTraditionInnovationScientists2015}. AI scientists carry the potential to address such problems and accelerate scientific progress.

\subsection{Where machine learning effort pays most}
Accelerating scientific discovery is among the most impactful applications of AI available to us; current AI systems for science, however, are usually built to demonstrate what the technology can do rather than to address what the science most needs, the technology-push regime of Section~\ref{sec:problem_selection}. Science differs from most other applications of machine learning. In most applications the prediction is the costly step and verification is cheap, so a better model is directly a better product. In science, verification by wet-lab experiment or high-precision simulation is far more expensive than the prediction, which changes what a useful system looks like. Systems that speed up experiments or simulations, reduce their error, or improve the correspondence between a fast proxy and the slow verification accepted as ground truth will be more impactful than systems that assemble existing community tools into agent loops and spend inference tokens navigating between them.

AlphaFold~\citep{jumperHighlyAccurateProtein2021} became impactful in large part by serving structural biologists as a first-pass structure determination tool. Machine-learned interatomic potentials~\citep{batznerE3EquivariantGraph2022, batatiaMACEHigherOrder2022} bring molecular dynamics close to density functional theory accuracy at a fraction of the cost, and universal potentials trained across the periodic table~\citep{dengCHGNetPretrainedUniversal2023} are now used as force fields throughout computational chemistry and materials science. \citet{kirkpatrickPushingFrontiersDensity2021} worked in the opposite direction, training a neural exchange-correlation functional that corrects long-standing systematic errors in density functional theory itself and releasing it through the open-source PySCF package, where quantum chemists use it as a functional rather than as an AI system. In each case, an AI team improved something the scientific community already depended on, and the improvement reaches every downstream user regardless of whether they use AI themselves. Progress toward AI scientists can also be positive even if the destination is never reached: while the need for human judgement and taste remains, progress toward autonomy will strengthen the AI co-scientists deployed for real problems.

\subsection{Conferences and accountability}
Agents4Science is already an example of a venue that explicitly invites research where AI systems are the primary contributors~\citep{bianchiExploringUseAI2026}. We have no mathematical certainty that a result found by an AI scientist will be correct and free of methodological error, so venues for public dissemination will need to weigh accountability at least as heavily as impact. We expect that, at least initially, the individual, group, or company that created the system will hold responsibility, and credit, for the results it produces.

A further consequence follows from velocity. Autonomous systems will generate candidate findings far faster than the present review system was built to absorb, and the rate of false positives entering the literature could rise. This is the same pressure our preregistration proposal in Section~\ref{sec:agenda} responds to, and it suggests that conferences and journals will need to become considerably more stringent about verifying claims than they are today. Conferences have already shifted away from being venues for purely academic feedback, increasingly drawing industry participants responsible for implementing scientific results in economic and social systems; that role may grow as trust in AI-discovered results increases. Conferences could also continue to serve as sites for exchanging ideas, particularly if proprietary AI scientists do not regularly publish their findings or deposit them in an accessible repository.

\subsection{Autonomous reviewers}
AI systems are likely to take on a substantial role in review well before they are capable of autonomous discovery. \citet{staracePaperBenchEvaluatingAIs2025} benchmark agentic systems on replicating results from AI publications, and the existence and design of that benchmark give us reason to believe the community will soon build models that perform such replication reliably. Such models would be valuable members of a review panel and one practical answer to the verification burden described above. Until an AI scientist capable of genuinely end-to-end discovery exists, however, we do not believe the final review decision should rest with an AI system. Part of what makes peer review work is the averaging of diverse and independent judgements, and the hypothesis hivemind experiment (Section~\ref{subsec:hivemind}) indicates that current AI systems are considerably more likely to converge. As argued under the alternative view that human scientists are also biased (Section~\ref{sec:alternative}), even a collection of AI systems that each individually outperform a human reviewer might not yield a panel with the diversity and self-correction that human review provides.

\subsection{De-skilling}
De-skilling of junior researchers is the inverse of the tacit-knowledge gap in training data. Section~\ref{sec:pretraining} argues that unwritten expertise is absent from published corpora because it is acquired by doing rather than by reading; a world in which research tasks are offloaded wholesale to AI scientists would, sooner or later, stop producing expert humans. Unlike specialised skills such as driving, where society may accept losing a human capability in exchange for the safety and convenience that autonomy provides, science and discovery are pursuits fundamental to human society. Human scientific capability must be preserved, because when an AI system proposes a paradigm shift or an unexpected result, humans must hold the underlying tacit knowledge to evaluate it critically rather than treat the system as an infallible oracle. When calculators were introduced, we neither ignored the technology nor removed arithmetic from mathematics education. Scientific education will need to adapt in a similar way: hands-on experimentation and reproduction of known results will need to become core pedagogical tools in research training, so that students build intuition and judgement by running experiments even where a model has already predicted the outcome accurately.

\end{document}

%% file: hivemind/numbers.tex
\newcommand{\hmNModels}{12}

\newcommand{\hmNSamples}{10}

\newcommand{\hmAiFourmatNovelIntraModel}{0.72}

\newcommand{\hmAiFourmatNovelCrossProvider}{0.64}

\newcommand{\hmAiFourmatNovelLevelCrossModel}{0.64}
\newcommand{\hmAiFourmatNovelLevelDifferentPaper}{0.39}
\newcommand{\hmAiFourmatNovelIdeasOneModel}{2.9}
\newcommand{\hmAiFourmatNovelIdeasTenModels}{3.6}
\newcommand{\hmAiFourmatNovelIdeasTenPapers}{6.5}

\newcommand{\hmPlosNovelIntraModel}{0.76}

\newcommand{\hmPlosNovelCrossProvider}{0.69}

\newcommand{\hmPlosNovelLevelCrossModel}{0.70}
\newcommand{\hmPlosNovelLevelDifferentPaper}{0.30}
\newcommand{\hmPlosNovelIdeasOneModel}{2.6}
\newcommand{\hmPlosNovelIdeasTenModels}{3.1}
\newcommand{\hmPlosNovelIdeasTenPapers}{7.4}

\newcommand{\hmAiFourmatNovelLenCorrIntra}{0.74}
\newcommand{\hmAiFourmatNovelLenCorrCross}{0.32}

\newcommand{\hmPlosNovelLenCorrIntra}{0.68}
\newcommand{\hmPlosNovelLenCorrCross}{0.27}
\newcommand{\hmAiFourmatNPapers}{50}

\newcommand{\hmPlosNPapers}{52}
\newcommand{\hmPlosNExcluded}{5}

%% file: hivemind/tables/models.tex
\resizebox{\linewidth}{!}{%
\begin{tabular}{lllccc}
\toprule
Model & Provider & Identifier & Open weights & AI4Mat & PLOS Biology \\
\midrule
Claude Haiku 4.5 & Anthropic & \texttt{anthropic/claude-haiku-4.5} &  & (a) & (b) \\
Claude Sonnet 4.5 & Anthropic & \texttt{anthropic/claude-sonnet-4.5} &  & (a) & (b) \\
Claude Sonnet 4.6 & Anthropic & \texttt{anthropic/claude-sonnet-4.6} &  & (a) & (b) \\
Gemma 4 31B & Google & \texttt{google/gemma-4-31b-it} & \checkmark & (c) & (c) \\
Gemini 3.1 Flash Lite & Google & \texttt{google/gemini-3.1-flash-lite} &  & (c) & (c) \\
Gemini 3.1 Pro Preview & Google & \texttt{google/gemini-3.1-pro-preview} &  & (c) & (c) \\
Kimi K2.6 & Moonshot AI & \texttt{moonshotai/kimi-k2.6} & \checkmark & (c) & (c) \\
Kimi K2.7 Code & Moonshot AI & \texttt{moonshotai/kimi-k2.7-code} & \checkmark & (c) & (c) \\
Kimi K3 & Moonshot AI & \texttt{moonshotai/kimi-k3} & \checkmark & (c) & (c) \\
GPT-5 Nano & OpenAI & \texttt{openai/gpt-5-nano} &  & (d) & (c) \\
GPT-5 Mini & OpenAI & \texttt{openai/gpt-5-mini} &  & (d) & (c) \\
GPT-5 & OpenAI & \texttt{openai/gpt-5} &  & (d) & (c) \\
\bottomrule
\end{tabular}}

\smallskip
\begin{minipage}{\linewidth}\footnotesize\raggedright
(a) Anthropic API; max\_tokens 4096; default temperature (1.0).\\
(b) OpenRouter (Anthropic protocol); max\_tokens 4096; default temperature.\\
(c) OpenRouter; max\_tokens 8192; reasoning effort \texttt{low}; default temperature.\\
(d) OpenAI API; temperature 1.0; default reasoning effort.\\
All models: 10 independent samples per paper and task; no seed; outputs embedded with \texttt{text-embedding-3-small}.
\end{minipage}

%% file: hivemind/tables/similarity_summary.tex
\begin{tabular}{llrccc}
\toprule
Dataset & Task & Papers & Intra-model & Intra-provider & Cross-provider \\
\midrule
AI4Mat & Recover underlying hypothesis & 50 & 0.86 {\scriptsize [0.85, 0.87]} & 0.81 {\scriptsize [0.80, 0.82]} & 0.77 {\scriptsize [0.76, 0.78]} \\
 & Generate novel hypothesis & 50 & 0.72 {\scriptsize [0.71, 0.73]} & 0.66 {\scriptsize [0.65, 0.67]} & 0.64 {\scriptsize [0.63, 0.65]} \\
\addlinespace
PLOS Biology & Recover underlying hypothesis & 52 & 0.88 {\scriptsize [0.87, 0.89]} & 0.83 {\scriptsize [0.82, 0.84]} & 0.81 {\scriptsize [0.80, 0.82]} \\
 & Generate novel hypothesis & 52 & 0.76 {\scriptsize [0.75, 0.77]} & 0.71 {\scriptsize [0.70, 0.72]} & 0.69 {\scriptsize [0.68, 0.70]} \\
\bottomrule
\end{tabular}

%% file: hivemind/tables/output_length.tex
\begin{tabular}{lrrrr}
\toprule
 & \multicolumn{2}{c}{AI4Mat} & \multicolumn{2}{c}{PLOS Biology} \\
\cmidrule(lr){2-3}\cmidrule(lr){4-5}
Model & Recover & Novel & Recover & Novel \\
\midrule
Claude Haiku 4.5 & 321 (58) & 382 (60) & 265 (54) & 344 (63) \\
Claude Sonnet 4.5 & 352 (73) & 503 (143) & 247 (48) & 462 (168) \\
Claude Sonnet 4.6 & 575 (94) & 640 (103) & 518 (95) & 610 (137) \\
Gemma 4 31B & 256 (44) & 297 (70) & 222 (47) & 247 (46) \\
Gemini 3.1 Flash Lite & 352 (74) & 362 (51) & 283 (61) & 335 (70) \\
Gemini 3.1 Pro Preview & 293 (48) & 333 (42) & 267 (63) & 292 (47) \\
Kimi K2.6 & 454 (100) & 346 (56) & 373 (93) & 307 (59) \\
Kimi K2.7 Code & 426 (88) & 328 (54) & 360 (80) & 296 (58) \\
Kimi K3 & 456 (95) & 456 (84) & 403 (93) & 413 (75) \\
GPT-5 Nano & 418 (95) & 403 (63) & 330 (82) & 334 (63) \\
GPT-5 Mini & 518 (92) & 416 (62) & 365 (68) & 320 (66) \\
GPT-5 & 481 (70) & 383 (65) & 399 (71) & 328 (59) \\
\midrule
$r$(length, intra-model) & 0.48 & 0.74 & 0.56 & 0.68 \\
$r$(length, cross-model) & 0.60 & 0.32 & 0.73 & 0.27 \\
\bottomrule
\end{tabular}